%% file: acl_latex.tex
\pdfoutput=1

\documentclass[11pt]{article}

\usepackage[final]{acl}

\usepackage{times}
\usepackage{latexsym}

\usepackage[T1]{fontenc}

\usepackage[utf8]{inputenc}

\usepackage{microtype}

\usepackage{inconsolata}

\usepackage{graphicx}

\usepackage{amssymb}
\usepackage{amsmath}
\usepackage{multirow} 
\usepackage{booktabs}
\usepackage{adjustbox}

\usepackage[bottom,hang,flushmargin]{footmisc}

\title{ConU: Conformal Uncertainty in Large Language Models with Correctness Coverage Guarantees}

\author{
 \textbf{Zhiyuan Wang\textsuperscript{1}},
 \textbf{Jinhao Duan\textsuperscript{2}},
 \textbf{Lu Cheng\textsuperscript{3}},
 \textbf{Yue Zhang\textsuperscript{2}},
 \textbf{Qingni Wang\textsuperscript{1}},
 \\
 \textbf{Xiaoshuang Shi\textsuperscript{1}\thanks{Corresponding to: Xiaoshuang Shi <xsshi2013@gmail.com>},}
 \textbf{Kaidi Xu\textsuperscript{2},}
 \textbf{Hengtao Shen\textsuperscript{1},}
 \textbf{Xiaofeng Zhu\textsuperscript{1}}
\\
\\
 \textsuperscript{1}School of Computer Science and Engineering, University of Electronic\\Science and Technology of China\\
 \textsuperscript{2}Department of Computer Science, Drexel University\\
 \textsuperscript{3}Department of Computer Science, University of Illinois Chicago
}

\begin{document}
\maketitle
\begin{abstract}
Uncertainty quantification (UQ) in natural language generation (NLG) tasks remains an open challenge, exacerbated by the closed-source nature of the latest large language models (LLMs). 
This study investigates applying conformal prediction (CP), which can transform any heuristic uncertainty notion into rigorous prediction sets, to black-box LLMs in open-ended NLG tasks. 
We introduce a novel uncertainty measure based on self-consistency theory, and then develop a conformal uncertainty criterion by integrating the uncertainty condition aligned with correctness into the CP algorithm. 
Empirical evaluations indicate that our uncertainty measure outperforms prior state-of-the-art methods. 
Furthermore, we achieve strict control over the correctness coverage rate utilizing 7 popular LLMs on 4 free-form NLG datasets, spanning general-purpose and medical scenarios.
Additionally, the calibrated prediction sets with small size further highlights the efficiency of our method in providing trustworthy guarantees for practical open-ended NLG applications. 
\end{abstract}

\input{section/introduction}
\input{section/related_work}

\input{section/method}

\input{section/experiment}
\input{section/conclusion}

\section*{Acknowledgments}
Zhiyuan Wang, Xiaoshuang Shi, and Xiaofeng Zhu were supported by the National Key Research $\&$ Development Program of China under Grant (No. 2022YFA1004100). 

\section*{Limitations}
Our approach has some limitations. 
We need to develop an uncertainty criterion to verify whether the correct answer has been sampled from the output space in real-world applications. 
Secondly, our findings are limited to the four datasets and future works will extend to other typical NLG tasks like document summarization. 
Finally, we will attempt to expand our conformal uncertainty criterion to non-exchangeability scenarios, aiming to establish a general criterion across different NLG tasks.

\bibliography{custom}

\appendix

\newpage
\input{section/appendix}

\end{document}

%% file: section/introduction.tex
\section{Introduction}
Despite advancements in various natural language generation (NLG) tasks~\cite{katz2024gpt, touvron2023llama, chen2023chatcot,duan2024reta,duan2024gtbench}, large language models (LLMs) are proven to hallucinate facts and confidently generate textual information that is not correct or grounded in reality~\cite{ji2023survey, manakul2023selfcheckgpt}. 
Factually incorrect answers can confuse and mislead users, resulting in erroneous conclusions and ultimately undermining the trustworthiness of LLMs-based high-stakes applications.

Uncertainty quantification (UQ) provides valuable insights into the reliability of model responses, facilitating risk assessment and hallucination detection~\cite{kadavath2022language, lin2022teaching}. 
However, it demands investigating black-box uncertainty measures with the proliferation of LLMs served via APIs~\cite{achiam2023gpt}, which only allows textual inputs and outputs. 
Conformal prediction (CP)~\cite{campos2024conformal, angelopoulos2021gentle, quach2023conformal,zhao2024conformalized} is known for providing a model-agnostic and statistically rigorous uncertainty estimation. 
CP was primarily employed in classification ~\cite{angelopoulos2021gentle} and regression tasks \cite{wang2024equal}. 
For NLG tasks, CP is first adapted to the multiple-choice question-answering (MCQA) setting, where the acceptable response is selected from a fixed set of options~\cite{kumar2023conformal,ye2024benchmarking}, limiting its applications in real-world open-ended NLG tasks. 
Conformal language modeling~\cite{quach2023conformal} relies on the model likelihoods and calibrates a stopping rule to sample prediction sets from the infinite output space until users are confident that the set covers at least one response satisfied. 
LofreeCP~\cite{su2024api} studies CP for API-only LLMs without logit access by leveraging uncertainty information from diverse sources.

Our study explores adapting CP for general NLG applications. 
The nonconformity score (NS) in CP serves as a criterion for calibrating prediction sets, which provide coverage guarantees by selecting a set of possible labels that satisfy the NS threshold~\cite{angelopoulos2021gentle}. 
Since typical logits-based NS may encounter miscalibration, we aim to integrate black-box UQ into the definition of NS, by closely aligning it with the uncertainty condition of the correct answers and devising a conformal uncertainty criterion, while it is more reliable to analyze the uncertainty within LLMs' true output space. 
Then, we employ the uncertainty criterion, concluded from a small amount of independent and identically distributed (i.i.d.) calibration data, to construct prediction sets by selecting generations sharing a similar uncertainty condition from the unbounded output space on test samples. Typically, there are two goals of CP: (1) the calibrated prediction set contains the correct answer with at least a user-specified probability; and (2) the average set size should be small, demonstrating the prediction efficiency of our method. 

The first challenge is UQ for black-box LLMs. 
Our solution is inspired by an intuitive observation: 
If a language model generates more semantically diverse outputs for the same prompt, the uncertainty is likely higher~\cite{su2024api, lin2023generating, xiong2023can}. 
Regardless of the model's capability to tackle the current problem, the confidence score that the model assigns to a generation can be represented by its frequency within the output space. 
We approximate the model's output distribution by sampling multiple answers to the same question. 
Then, we perform semantic clustering on the sampled generations, and propose to measure the uncertainty of each generation by combining two factors: the frequency of occurrence of the semantic meaning it conveys, and the consistency between its semantic and other semantic clusters augmented by their individual frequency. 

Based on the measure, we define the NS as the uncertainty of the generation. To this end, the generation meets the correctness criterion and is semantically most similar to the reference answer in the calibration set. 
We then calculate the quantile $\hat{q}$ of NSs for all calibration samples, based on the user-specified upper bound of error rate $\alpha$. 
Next, we utilize the conformal uncertainty criterion (i.e., the uncertainty threshold $\hat{q}$) to construct a prediction set for each test sample by selecting generations that satisfy the uncertainty conditions strictly associated with correctness from the candidate generations. 
Additionally, for black-box UQ, we propose employing the most frequent generation or semantic (i.e., the model's most confident answer) as a more trustworthy reference object for the query and leveraging it to measure the overall uncertainty of the current UQ process. 
We term this measure \textit{ConU}, as it employs the same approach as the conformal uncertainty criterion. 

Extensive experimental results exhibit that \textit{ConU} generally outperforms prior state-of-the-art methods and verify the strict correctness coverage guarantees. Specifically, the prediction sets calibrated by the conformal uncertainty criterion always encompass the correct answers under various user-specified error rates. 
Furthermore, the average prediction set size is small, highlighting the prediction efficiency of our approach. 
To our knowledge, this is the first method in the literature to strictly link the NS with the uncertainty condition aligned with correctness via black-box UQ, thereby developing a more robust conformal uncertainty criterion, which provides rigorous correctness coverage guarantees in practical open-ended NLG tasks, and its unique inspiration in benchmarking UQ in LLMs through CP generates independent interest\footnote{Our code is available at \href{https://github.com/Zhiyuan-GG/Conformal-Uncertainty-Criterion/tree/main}{https://github.com/Zhiyuan-GG/Conformal-Uncertainty-Criterion/tree/main}}. 

In summary, our major contributions are listed as follows:

\begin{itemize}
    \item We propose a sampling-based black-box uncertainty measure, termed as \textit{ConU}, utilizing self-consistency in open-ended NLG tasks, facilitating trustworthy decision-making. 
    \item We devise a conformal uncertainty criterion by strictly aligning the NS with the uncertainty condition of acceptable answers, and achieve rigorous correctness coverage with at least a user-specified probability, thereby providing robust guarantees under various error rates in practical open-ended NLG applications. 
    \item We conduct selective prediction leveraging the calibrated prediction sets and obtain promising improvements in model accuracy without requiring additional task-specific fine-tuning or architectural modifications. 
\end{itemize}

%% file: section/related_work.tex
\section{Related Work}
\subsection{Uncertainty Quantification in LLMs}
Prior work on UQ in LLMs predominantly focuses on white-box information like token-likelihoods or  embeddings~\cite{da2024llm,kuhn2023semantic, duan2023shifting, wang2024word}, internal state or activations~\cite{yin2024characterizing, chen2024inside}, model fine-tuning~\cite{tian2023just}. 
These methods can encounter poor calibration and require substantial computational resources.
Additionally, researchers lack white-box access to the internal information of LLMs served via APIs. 
These restrictions demand black-box measures for general UQ in LLMs generations. 

Recent work~\cite{lin2023generating} develops several sampling-based uncertainty measures, which can be applied to black-box LLMs by leveraging semantic similarity along with dispersion. 
Our study follows the sampling setting and proposes to employ the most frequent generation as the reference object to measure the overall uncertainty based on the self-consistency theory~\cite{wang2022self}. 

\subsection{Conformal Prediction in LLMs}\label{sec: related cp work}
CP~\cite{angelopoulos2021gentle, quach2023conformal, campos2024conformal} has emerged as a theoretically sound and practically useful way to guarantee ground-truth coverage with the aid of a small amount of exchangeable samples for calibration. 
CP in classification tasks defines the NS, which is correlated with the ground-truth label, obtains the quantile, $\hat{q}$, of NSs for all calibration samples based on a user-specified upper bound of the error rate $\alpha$, and utilizes $\hat{q}$ as a threshold to select possible labels on test samples, thereby establishing prediction sets that guarantee ground truth coverage with at least the probability of $1-\alpha$. 

Recently, researchers have attempted to apply CP to LLMs for principled UQ. 
The work~\cite{mohri2024language} achieves conformal factuality guarantees by progressively making generations less specific and establishing their corresponding entailment sets until correct answers are encompassed. 
For correctness coverage, two studies~\cite{kumar2023conformal, ye2024benchmarking} follow CP in classification tasks and convert NLG tasks into MCQA settings. 
For open-ended NLG, based on the output token sequence logits, the study~\cite{quach2023conformal} develops a stopping rule to sample generations until users are confident that a correct answer is covered in QA tasks, which can be impractical for API-only LLMs. 
LofreeCP~\cite{su2024api} leverages uncertainty information to construct prediction sets that achieve correctness coverage. 

This paper focuses on more practical scenarios of black-box LLMs in open-ended NLG tasks. 
Differing from LofreeCP, we strictly connect the NS with the uncertainty condition aligned with correctness via black-box UQ, which concludes a more robust conformal uncertainty criterion to calibrate prediction sets with rigorous correctness coverage guarantees under various error rates despite the complexity of the model or datasets.

%% file: section/method.tex
\section{Method}
Our method investigates two key issues: (1) how to estimate the uncertainty in black-box LLMs when we can only access the output texts; and (2) how to provide rigorous guarantees on the error rate in open-ended NLG tasks. 
We first devise a black-box uncertainty measure grounded in self-consistency to provide the trustworthiness notion of model responses. 
Furthermore, we utilize the split CP technique to convert the heuristic approximation into a statistically rigorous one, thereby ensuring a more robust and systematic assessment of uncertainty. 

\subsection{Preliminaries}
Following the analysis of black-box LLMs in prior work~\cite{xiong2023can, lin2023generating, manakul2023selfcheckgpt}, conditioned on each prompt (or question) $x_i$, we employ the most likely generation $\hat{y}_i$ for correctness evaluation. 
Additionally, we sample a set of $M$ candidate generations $\left\{ \hat{y}_m^{(i)} \right\}_{m=1}^M$ from the model's output space for black-box UQ and the derivation of conformal uncertainty criterion. 
We denote the reference answer to $x_i$ as $y_i^*$. 

\subsection{Uncertainty Quantification}
For each data point, we first cluster semantics in the $M$ sampled generations and obtain $K$ non-repeated semantics. We denote the number of generations sharing the $k$-th semantic as $V_k$ (i.e., $\textstyle\sum_{k=1}^{K}V_k=M$) and any one generation in this cluster as $\hat{y}_k^{(i)}$.

Building on earlier approaches that utilize self-consistency~\cite{wang2022self, su2024api, yadkori2024mitigating} as a reliable measure of confidence, we employ the frequency of the $k$-th semantic as its proxy for reliability: $\mathcal{F}\left(\hat{y}_k^{(i)}\right)=\frac{V_k}{M}$.
Then, we define the uncertainty score of each candidate generation in $\left\{ \hat{y}_m^{(i)} \right\}_{m=1}^M$ as 
\begin{equation}\label{eq: uncertainty score}
\begin{split}
    \mathcal{U}\left( \hat{y}_m^{(i)} \right) = &1 - \lambda \cdot \mathcal{F}\left(\hat{y}_m^{(i)}\right) - \left(1-\lambda\right) \cdot\\ 
    &\frac{1}{K} \displaystyle\sum_{k=1}^{K}\mathcal{S}\left ( \hat{y}_m^{(i)}, \hat{y}_k^{(i)} \right )\mathcal{F}\left(\hat{y}_k^{(i)}\right),
\end{split}
\end{equation}
where $\mathcal{F}\left(\hat{y}_m^{(i)}\right)$ refers to the frequency of the semantic that $\hat{y}_m^{(i)}$ conveys, and $\mathcal{S}\left( \cdot,\cdot \right)$ measures the semantic similarity between two generations utilizing a \textit{cross-encoder} model~\cite{reimers2019sentence}. $\mathcal{F}\left(\hat{y}_k^{(i)}\right)$ is to augment the persuasiveness of the similarity score associated with $\hat{y}_k^{(i)}$.

To measure the model uncertainty, we select any one generation in the largest semantic cluster to be the most trustworthy generation in the $M$ sampled generations and denote it as $\hat{y}_{mst}^{{i}}$. 
Then, we define the uncertainty score of the $i$-th query-response process as
\begin{equation}
\begin{split}
    &\mathcal{U}\left(\left\{ \hat{y}_m^{(i)} \right\}_{m=1}^M|x_i\right) =1- \lambda \cdot \mathcal{F}\left(\hat{y}_{mst}^{{i}}\right) -\\
    &\left(1-\lambda\right) \cdot \frac{1}{K} \displaystyle\sum_{k=1}^{K}\mathcal{S}\left (\hat{y}_{mst}^{{i}}, \hat{y}_k^{(i)} \right )\mathcal{F}\left(\hat{y}_k^{(i)}\right).
\end{split}
\end{equation}
Intuitively, the most frequent semantic within the candidate generations represents the model's most confident answer to the current problem. 
Even though the reference semantic may not necessarily be the correct one, we can measure the degree of the model's uncertainty by calculating the confidence level of that semantic as well as the deviation between it and other semantics.  

Since Eq.~\eqref{eq: uncertainty score} can quantify the uncertainty of each candidate generation, we attempt to develop an uncertainty criterion to search for the correct answers within the unfixed output space of the LLM.

\subsection{Conformal Correctness Coverage}\label{sec: conformal prediction}
Following the fundamental requirement in split CP~\cite{angelopoulos2021gentle}, we randomly employ $N$ samples to construct the calibration data set $\left\{ \left( x_i, y_i^* \right) \right\}_{i=1}^N$, and for each calibration sample we demand that at least one sampled generation $\hat{y}_j^{(i)}$ in $\left\{ \hat{y}_m^{(i)} \right\}_{m=1}^M$ meets the correctness criterion. 
Our objective of \textbf{conformal correctness coverage} is by concluding the uncertainty criterion that is closely linked with correctness on $\left\{ \left( x_i, y_i^* \right) \right\}_{i=1}^N$, we can calibrate an uncertainty (prediction) set $\mathcal{P}\left(x_{test}\right)$ for the test prompt $x_{test}$ by selecting generations that meet the common uncertainty condition, 
and the set can guarantee correctness coverage under various user-specificed error rates. 
Here, we approximate the prediction region of $x_{test}$ to the $M$ candidate generations $\left\{ \hat{y}_m^{(test)} \right\}_{m=1}^M$.

\textit{Assumptions: (1) There is at least one candidate generation in $\left\{ \hat{y}_m^{(test)} \right\}_{m=1}^M$ meeting the correctness criterion; (2) Samples in the calibration and test data sets are exchangeable.}

As the sampled set $\left\{ \hat{y}_m^{(test)} \right\}_{m=1}^M$ is a subset of the prediction region,  which is impossible to enumerate, we can simplify it by stating that there is at least one correct answer in $\left\{ \hat{y}_m^{(test)} \right\}_{m=1}^M$. Exchangeability is the fundamental assumption of CP~\cite{angelopoulos2021gentle}. 
We provide the explanation for Assumption (1) in Appendix~\ref{sec: assumption1}.

Based on the uncertainty measure described as Eq.~\eqref{eq: uncertainty score}, we define the NS of the $i$-th calibration sample as
\begin{equation}
\begin{split}
    &r_i = r\left ( x_i, y_i^* \right )=\\
    & \mathcal{U}\left( {\arg\max}_{\hat{y}_j^{(i)}} \mathcal{S}\left( \hat{y}_j^{(i)}, y_i^* \right) \mathcal{E}\left( \hat{y}_j^{(i)}, y_i^* \right)\right),
\end{split}\label{eq: NS}
\end{equation}
where $\mathcal{E}\left(\cdot,\cdot\right)$ is the indicator function determining whether the two sentences share equivalent semantics, i.e., $\mathcal{E}\left ( \hat{y}_j^{(i)},y_i^* \right )=1$ indicates that $\hat{y}_j^{(i)}$ is semantically equivalent to $y_i^*$, and $\mathcal{E}\left ( \hat{y}_j^{(i)},y_i^* \right )=0$ denotes it does not. 
That is, the NS, $r\left ( x_i, y_i^* \right )$ represents the uncertainty condition of the candidate generation $\hat{y}_j^{(i)}$, which has the highest similarity score with the reference answer $y_i^*$ in generations that are semantically equivalent to $y_i^*$.
The criterion for determining semantic equivalence here is the same as that for correctness evaluation (i.e., $\hat{y}_j^{(i)}$ is correct according to $y_i^*$ if $\mathcal{E}\left ( \hat{y}_j^{(i)},y_i^* \right )=1$).

It is worth emphasizing that we strictly align the NSs with the uncertainty conditions of correct answers within the fresh calibration set, concluding an honest insight into the model's performance, which is crucial for robust correctness coverage guarantees in new test samples. 

Following prior work~\cite{angelopoulos2021gentle, quach2023conformal, campos2024conformal}, we sort $\left\{ r_i \right\}_{i=1}^N$ ($\left \{r_1 \leq \cdots \leq r_N \right \}$) and calculate the $\frac{\left \lceil \left ( N+1\right )\left ( 1-\alpha \right )\right \rceil}{N}$ quantile of NSs for all calibration data to develop the conformal uncertainty criterion 
\begin{equation}
\begin{split}
    &\hat{q} =\\ 
    &\inf \left \{ q:\frac{\left | \left \{ i:r_i \leq  q\right \}\right |}{N} \geq \frac{\left \lceil \left ( N+1\right )\left ( 1-\alpha \right )\right \rceil}{N} \right \}\\
    &={r}_{\left \lceil \left ( N+1\right )\left ( 1-\alpha \right )\right \rceil},
\end{split}\label{eq: q_hat}
\end{equation}
where $\alpha$ is the upper bound of the error rate.

\input{tables/uncertainty_quantification_ss}

As for each test sample, we construct the prediction set following 
\begin{equation}\label{eq: prediction set}
\begin{split}
    \mathcal{P}\left(x_{test}\right)=\left \{ \hat{y}_j^{(test)} : r\left( x_{test}, \hat{y}_j^{(test)} \right) \leq \hat{q} \right \}.
\end{split}
\end{equation} 
It is evident that the most semantically similar generation to $\hat{y}_j^{(test)}$ in $\left\{ \hat{y}_m^{(test)} \right\}_{m=1}^M$ is itself, and we obtain $r\left( x_{test}, \hat{y}_j^{(test)} \right) = \mathcal{U}\left( \hat{y}_j^{(test)} \right)$. 
Recall the assumption that $\left\{ \hat{y}_m^{(test)} \right\}_{m=1}^M$ contains at least one correct generation (i.e., $y_{test}^* \in \left\{ \hat{y}_m^{(test)} \right\}_{m=1}^M$),  then the event $\left\{ y_{test}^* \in  \mathcal{P}\left(x_{test}\right) \right\}$ is equivalent to $\left\{ r_{test}=r\left( x_{test}, y_{test}^* \right) \leq \hat{q} \right\}$. 

Since the calibration and test samples $\left(x_1, y_1^*\right)$, ..., $\left(x_N, y_N^*\right)$, $\left(x_{test}, y_{test}^*\right)$ are exchangeable, we have $P\left( r_{test} \leq r_i \right)=\frac{i}{N+1}$. 
Then we conclude 
\begin{equation}\label{eq: lower bound}
\begin{split}
    P\left( y_{test}^* \in  \mathcal{P}\left(x_{test}\right) \right) &= P\left( r_{test} \leq {r}_{\left \lceil \left ( N+1\right )\left ( 1-\alpha \right )\right \rceil} \right) \\
    &= \frac{\left \lceil \left ( N+1\right )\left ( 1-\alpha \right )\right \rceil}{N+1}\\
    &\geq 1-\alpha,
\end{split}
\end{equation}
and obtain the user-specified lower bound (i.e., $1-\alpha$) of the correctness coverage rate guaranteed by these calibrated prediction sets.

%% file: tables/uncertainty_quantification_ss.tex
\begin{table*}[t!]
\centering
\caption{Performance comparison (AUROC) of uncertainty quantification across our proposed method and 8 baseline approaches, evaluated on 5 instruction-tuned LLMs over 4 open-ended NLG datasets. The correctness criterion is based on the sentence similarity measured by the DistillRoBERTa model with a threshold of 0.7. The best UQ methods are in \textbf{bold} and the second-best one is \underline{underscored}.}
\adjustbox{max width=\linewidth}{
    \begin{tabular}{c|c|cccc|cccc|c} 
        \toprule
         \multirow{2}{*}{Dataset} & \multirow{2}{*}{LLMs} & \multicolumn{4}{c|}{White-box} & \multicolumn{5}{c}{Black-box} \\
         \cline{3-11}
         \multirow{2}{*}{} & \multirow{2}{*}{} & \textit{PE} & \textit{LNPE} & \textit{SE} & \textit{SAR} & \textit{LS} & \textit{NumSet} & \textit{Ecc} & \textit{Deg} & \textit{ConU}\\
         
        \midrule
        
        \multirow{5}{*}{TriviaQA} & LLaMA-2-7B-Chat & 0.6587 & 0.6459 & 0.7495 & 0.7876 & 0.5571 & 0.7763 & 0.7839 & \underline{0.8103} & \textbf{0.8198}\\
        \multirow{5}{*}{} & Mistral-7B-Instruct-v0.3 & 0.6620 & 0.5968 & 0.7845 & 0.8306 & 0.5969 & 0.8491 & \underline{0.8596} & \underline{0.8596} & \textbf{0.8671}\\
        \multirow{5}{*}{} & LLaMA-3-8B-Instruct & 0.7247 & 0.6465 & 0.7934 & \underline{0.8271} & 0.4661 & 0.8201 & 0.7404 & 0.8246 & \textbf{0.8275}\\
        \multirow{5}{*}{} & Vicuna-13B-v1.5 & 0.5553 & 0.5543 & 0.7568 & 0.7207 & 0.5734 & 0.7629 & 0.6578 & \underline{0.7858} & \textbf{0.7926}\\
        \multirow{5}{*}{} & LLaMA-2-13B-Chat & 0.6065 & 0.5614 & 0.7624 & 0.7757 & 0.6121 & 0.7885 & \underline{0.8035} & \underline{0.8035} & \textbf{0.8048}\\

        \midrule
        \multicolumn{2}{c|}{Average} & 0.6414 & 0.6010 & 0.7693 & 0.7883 & 0.5611 & 0.7994 & 0.7690 & \underline{0.8167} & \textbf{0.8224}\\
        \midrule

        \multirow{5}{*}{CoQA} & LLaMA-2-7B-Chat & 0.6236 & 0.5618 & 0.7120 & 0.7372 & 0.5403 & 0.7309 & 0.6769 & \textbf{0.7613} & \underline{0.7600}\\
        \multirow{5}{*}{} & Mistral-7B-Instruct-v0.3 & 0.6746 & 0.5795 & 0.7062 & 0.7551 & 0.5799 & 0.7481 & 0.6931 & \underline{0.7645} & \textbf{0.7652}\\
        \multirow{5}{*}{} & LLaMA-3-8B-Instruct & 0.7495 & 0.6531 & 0.7652 & \textbf{0.7902} & 0.4532 & 0.7400 & 0.7288 & \underline{0.7763} & 0.7702\\
        \multirow{5}{*}{} & Vicuna-13B-v1.5 & 0.5928 & 0.5565 & \underline{0.7110} & 0.6984 & 0.4965 & 0.6832 & 0.6679 & \textbf{0.7191} & 0.7106\\
        \multirow{5}{*}{} & LLaMA-2-13B-Chat & 0.6203 & 0.5634 & 0.7039 & 0.7427 & 0.5534 & 0.7230 & 0.6805 & \underline{0.7546} & \textbf{0.7591}\\

        \midrule
        \multicolumn{2}{c|}{Average} & 0.6522 & 0.5829 & 0.7197 & 0.7472 & 0.5247 & 0.7250 & 0.6894 & \textbf{0.7552} & \underline{0.7530}\\
        \midrule

        \multirow{5}{*}{MedQA} & LLaMA-2-7B-Chat & 0.4888 & 0.4925 & 0.5341 & 0.5862 & 0.5599 & 0.5933 & 0.5511 & \underline{0.6064} & \textbf{0.6120}\\
        \multirow{5}{*}{} & Mistral-7B-Instruct-v0.3 & 0.4613 & 0.4639 & 0.5091 & 0.6397 & 0.5520 & 0.6282 & 0.6562 & \underline{0.6660} & \textbf{0.6789}\\
        \multirow{5}{*}{} & LLaMA-3-8B-Instruct & 0.5854 & 0.5781 & 0.6508 & \underline{0.7167} & 0.4522 & 0.7093 & 0.6142 & 0.7159 & \textbf{0.7196}\\
        \multirow{5}{*}{} & Vicuna-13B-v1.5 & 0.4970 & 0.4922 & 0.5523 & 0.5854 & 0.5479 & 0.5926 & 0.5383 & \underline{0.6261} & \textbf{0.6360}\\
        \multirow{5}{*}{} & LLaMA-2-13B-Chat & 0.4618 & 0.4647 & 0.5277 & 0.5792 & 0.5734 & 0.6041 & 0.5743 & \underline{0.6070} & \textbf{0.6153}\\

        \midrule
        \multicolumn{2}{c|}{Average} & 0.4989 & 0.4983 & 0.5548 & 0.6214 & 0.5371 & 0.6255 & 0.5868 & \underline{0.6443} & \textbf{0.6524}\\
        \midrule

        \multirow{5}{*}{MedMCQA} & LLaMA-2-7B-Chat & 0.4774 & 0.4848 & 0.5221 & 0.5883 & 0.5531 & \underline{0.6171} & 0.5165 & 0.5983 & \textbf{0.6330}\\
        \multirow{5}{*}{} & Mistral-7B-Instruct-v0.3 & 0.4971 & 0.4989 & 0.5491 & 0.6944 & 0.5103 & 0.7084 & 0.7170 & \underline{0.7173} & \textbf{0.7413}\\
        \multirow{5}{*}{} & LLaMA-3-8B-Instruct & 0.5414 & 0.5395 & 0.6244 & 0.6940 & 0.4817 & 0.6992 & 0.5952 & \underline{0.6993} & \textbf{0.7098}\\
        \multirow{5}{*}{} & Vicuna-13B-v1.5 & 0.4614 & 0.4815 & 0.5550 & 0.5509 & 0.5377 & 0.5891 & 0.5135 & \underline{0.6221} & \textbf{0.6448}\\
        \multirow{5}{*}{} & LLaMA-2-13B-Chat & 0.4547 & 0.4712 & 0.5385 & 0.5701 & 0.5711 & \underline{0.6378} & 0.6188 & 0.6188 & \textbf{0.6414}\\

        \midrule
        \multicolumn{2}{c|}{Average} & 0.4864 & 0.4952 & 0.5578 & 0.6195 & 0.5308 & 0.6503 & 0.5922 & \underline{0.6511} & \textbf{0.6741}\\
        
        \bottomrule
    \end{tabular}
}
\label{tb: auroc ss}
\end{table*}

%% file: section/experiment.tex
\section{Evaluations}
\subsection{Experimental Set-up}
\paragraph{Baselines.}
We consider 8 baseline methods, including 4 white-box methods: Predictive Entropy (\textit{PE})~\cite{kadavath2022language}, Length-normalized Predictive Entropy (\textit{LNPE})~\cite{malinin2020uncertainty}, Semantic Entropy (\textit{SE})~\cite{kuhn2023semantic}, and Shift Attention to Relevance (\textit{SAR})~\cite{duan2023shifting}, and 4 black-box approaches: Lexical Similarity (\textit{LS})~\cite{lin2022towards} and Number of Semantic Sets (\textit{NumSet})~\cite{kuhn2023semantic, lin2023generating}. Moreover, we also include the most recent state-of-the-art uncertainty quantification methods, Degree Matrix (\textit{Deg})~\cite{lin2023generating}, and Eccentricity (\textit{Ecc})~\cite{lin2023generating}. 
More details of baseline methods can be found in Appendix~\ref{sec: ap baselines}.

\paragraph{Base LLMs.}
We conduct empirical evaluations on 7 LLMs encompassing various sizes and architectures for comprehensive analysis, including GPT-3.5-turbo served by OpenAI\cite{gpt3.5}, LLaMA-2-7B-Chat~\cite{touvron2023llama2}, Mistral-7B-Instruct-v0.3~\cite{jiang2023mistral}, Llama-3-8B-Instruct~\cite{llama3modelcard}, Vicuna-13B-v1.5~\cite{zheng2023judging}, LLaMA-2-13B-Chat~\cite{touvron2023llama2}, LLaMA-3-70B-Instruct~\cite{llama3modelcard}. 
We utilize the default generation configs and checkpoints provided by the HuggingFace platform\footnote{\href{https://huggingface.co/models}{https://huggingface.co/models}} for all open-source LLMs.

\begin{figure}[!t]
\centerline{\includegraphics[width=\columnwidth]{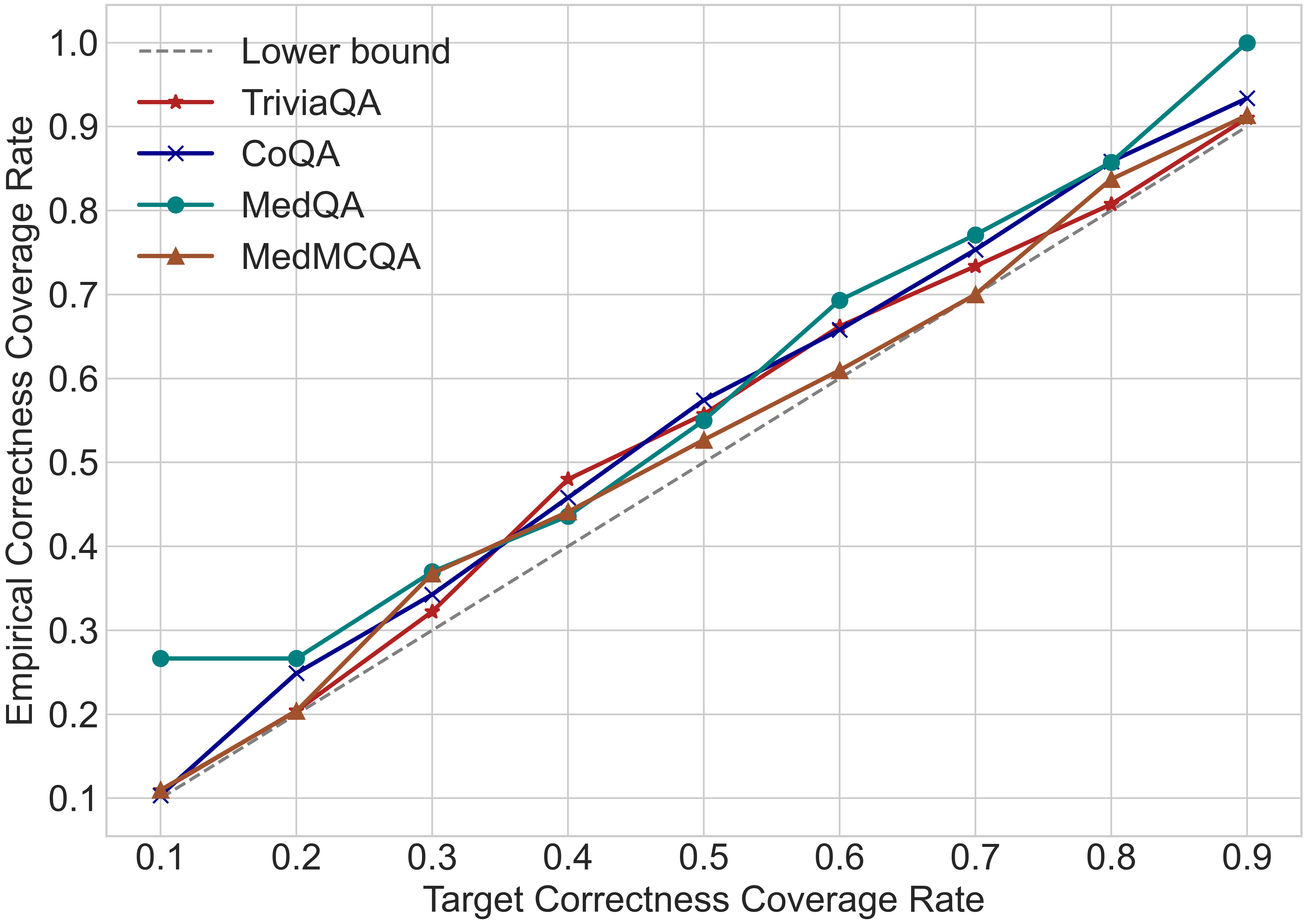}}
\caption{Target vs. empirical correctness coverage rate.\\We test the 4 datasets utilizing the LLaMA-2-7B-Chat model as the generator. Empirically, we achieve strict control over the coverage of correct answers by calibrating prediction sets on 4 free-form QA datasets.}\label{fig: conformal correctness coverage}
\end{figure}

\paragraph{Datasets.}
We evaluate the performance of \textit{ConU} and verify the correctness coverage guarantees on 4 free-form NLG datasets, including CoQA~\cite{reddy2019coqa} for conversational QA task, TriviaQA~\cite{joshi2017triviaqa} for reading comprehension, MedQA~\cite{jin2021disease} for solving medical problems, and MedMCQA~\cite{pal2022medmcqa} for medical entrance exam questions. 
More details of datasets can be found in Appendix~\ref{sec: ap datasets}.

\paragraph{Evaluation Metric.}
Following prior work~\cite{duan2023shifting, wang2024word}, we evaluate the performance of UQ by treating it as the problem of predicting whether to trust a generation given the prompt, and utilize the Area Under the Receiver Operating Characteristic Curve (AUROC) which gauges if the uncertainty scores can effectively distinguish between correct and incorrect generations. 
To verify if the correctness coverage is strictly guaranteed, we evaluate the coverage rate under various user-specified error rates. 
We also report the average prediction set size to evaluate the prediction efficiency and practicality of our approach. 

\paragraph{Correctness and Equivalence Metric.}
We utilize sentence similarity~\cite{duan2023shifting} as the metric for correctness and equivalence evaluation. 
We employ the cross-encoder model~\cite{reimers2019sentence} with DistillRoBERTa~\cite{sanh2019distilbert} serving as the backbone to measure the semantic similarity score between the most likely generation and reference answer and set a strict correctness threshold of 0.7. 

\input{tables/coverage_rate}

\input{tables/prediction_set_size}

\paragraph{Hyperparameters.}
We randomly sample 5 answers to each question for UQ and 10 candidate generations for verification of correctness coverage guarantees. 
We leverage beam search for the most likely generations for correctness evaluation and multinominal sampling for candidate generations~\cite{duan2023shifting}. 
The max length of each generation is set to 128 tokens. 
The temperature of generation is set to 1.0. 
The coefficient $\lambda$ introduced in Eq.~\eqref{eq: uncertainty score} is set to 0.5. 
The ratio of calibration and test set is set to 1:10 by default. 

\subsection{UQ in Black-Box LLMs}
As defined in failure prediction~\cite{xiong2023can} which evaluates whether the uncertainty score can effectively distinguish between correct and incorrect generations, an effective measure should assign higher uncertainty to incorrect generations and lower to correct ones. 
We compare our approach with state-of-the-art methods utilizing AUROC. 
Experimental results are summarized in Table~\ref{tb: auroc ss}. 
Generally, our method outperforms baseline methods in most of the settings. 
For instance, our method consistently beat 8 baseline methods on the TriviaQA datasets. 
It is worth noting that our method outperforms other methods by at most 2.4$\%$ AUROC on the MedMCQA dataset and 1.29$\%$ AUROC on the MedQA, which indicates the potential impacts of our methods on real-world high-stakes NLG applications. 
We will discuss the impact of the number of sampled generations on UQ in Section~\ref{sec: ablation}.


\begin{figure}[!t]
\centerline{\includegraphics[width=\columnwidth]{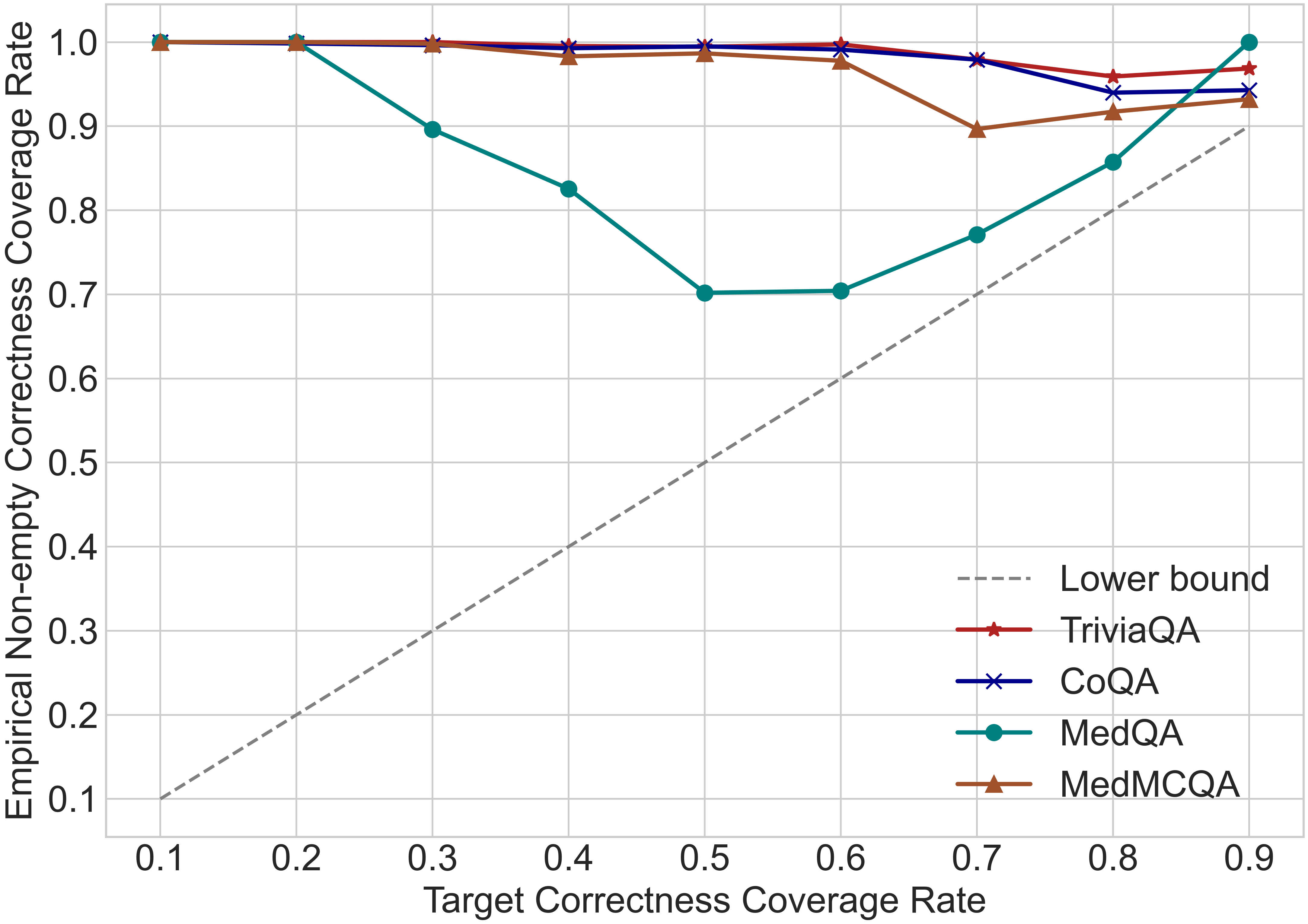}}
\caption{Target correctness coverage rate vs. empirical correctness coverage rate on non-empty prediction sets. We test the 4 datasets utilizing the LLaMA-2-7B-Chat model. We can almost obtain absolute coverage of correct answers in non-empty calibrated prediction sets even at a strict user-accepted error rate.}
\label{fig: non-empty conformal correctness coverage}
\end{figure}

\subsection{Conformal Correctness Coverage}
In this section, we verify that the calibrated prediction sets constructed following Eq.~\eqref{eq: prediction set} indeed achieve rigorous correctness coverage guarantees under various user-specified error rates as described in Eq.~\eqref{eq: lower bound}. Then we explore the utility of prediction sets and conduct selective prediction based on our proposed uncertainty measure. 

\paragraph{Empirical Coverage Guarantees.}
To guarantee the derived lower bound of correctness coverage rate in practice, we randomly split the four datasets at a ratio of 1:10, employing the respective portions as the calibration and test set. 
We utilize the calibration set to derive the conformal uncertainty criterion specified by the upper bound of the error rate.
Then, we measure the correctness coverage rate on the test set and plot the results on four datasets in Figure~\ref{fig: conformal correctness coverage}. 
It is evident that we achieve strict control of the correctness coverage rate under various error rates. 
The verification on other models can be found in Appendix~\ref{sec: ap correctness coverage guarantees}. 

Following the study~\cite{ye2024benchmarking}, we set the error rate $\alpha$ to 0.1 and test the coverage rate on 4 datasets utilizing 7 LLMs with various scales. 
As is exhibited in Table~\ref{tb: cr}, the coverage rate is at least $90\%$, indicating that the requirement of correctness coverage guarantees is satisfied. 
It is worth noting that prior work~\cite{ye2024benchmarking, kumar2023conformal} selects the possible option from the fixed choices while we characterize the unbound answer distribution by sampling and utilize our devised conformal uncertainty criterion to search for the correct answer, which is more practical. 

\input{tables/accuracy_calibration}

We also evaluate the prediction efficiency of the conformal uncertainty criterion utilizing the average size of these calibrated prediction sets, which is the primary metric for CP~\cite{angelopoulos2021gentle}. 
Table~\ref{tb: prediction set size} demonstrates that the average size of prediction sets calibrated by our method remains very small across the 4 datasets. 
For instance, the average set size is 1.03 on the LLaMa-3-70B-Instruct model in the TriviaQA task, indicating that we can almost directly identify the correct answers through these calibrated prediction sets. 

We boldly expect that as long as the language model has the capability to solve the current problem, despite the unfixed answer distribution, we can always find the correct generation by performing black-box UQ on each sampled answer and searching for answers meeting the conformal uncertainty criterion, and then limit the selection region to the calibrated prediction set for post-processing.
 

\paragraph{Utility of Calibrated Prediction Sets.}
Since for some test samples, all the candidate generations can be filtered out by the conformal uncertainty criterion, we explore the utility of non-empty prediction sets in practice. 
Figure~\ref{fig: non-empty conformal correctness coverage} exhibits that the prediction sets achieve promising correctness coverage rate, raising to 100$\%$ as the accepted error rate increases. 
In the MedQA dataset, while the error rate is set to 0.1, we almost achieve absolute correctness coverage guarantees, indicating that, without reference answers provided in real-world high-stakes situations, we can ensure that the small reference range we have established contains the correct answer for posterior selection, and then high-uncertainty problems will be handed over to experts, which aligns with the selective prediction and abstention criterion.

Based on the proposed uncertainty measure, we conduct post-processing to select the generation with the lowest uncertainty score from each calibrated prediction set and evaluate the total selective accuracy. 
It is worth noting that the performance depends on the quality of the uncertainty measure. 
Results are summarized in Table~\ref{tb: acc}. 
Through posterior selection, we obtain promising accuracy improvement despite several empty prediction sets.

\subsection{Ablation Studies}\label{sec: ablation}
\begin{figure}[!t]
\centerline{\includegraphics[width=\columnwidth]{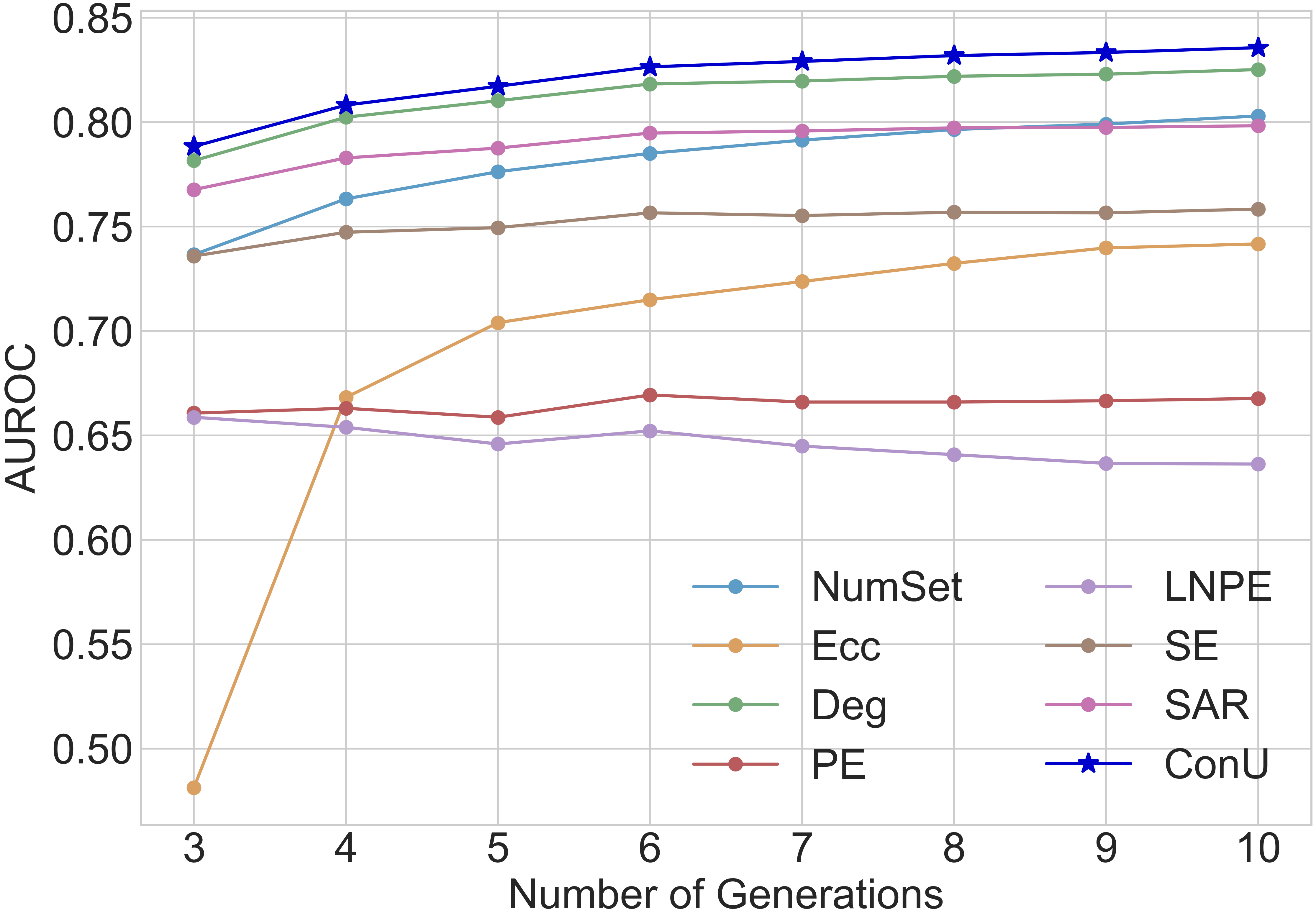}}
\caption{The performance of UQ over various numbers of generations. Results are obtained from the LLaMA-3-8B-Instruct model on the TriviaQA dataset. Our method consistently surpasses 7 baseline methods. }
\label{fig: multi-generation auroc}
\end{figure}

Considering that these sampling-based methods integrate multiple generations within the candidate set, We investigate the effects of the number of sampled generations (i.e., $M$) on the performance of UQ. 
As illustrated in Figure~\ref{fig: multi-generation auroc}, our uncertainty measure consistently outperforms the baseline approaches, and its performance can be further boosted by incorporating more generations. 
While employing just 4 generations, our method is able to achieve the highest AUROC of 0.8082, demonstrating its generation-efficient nature.  

\begin{figure}[!t]
\centerline{\includegraphics[width=\columnwidth]{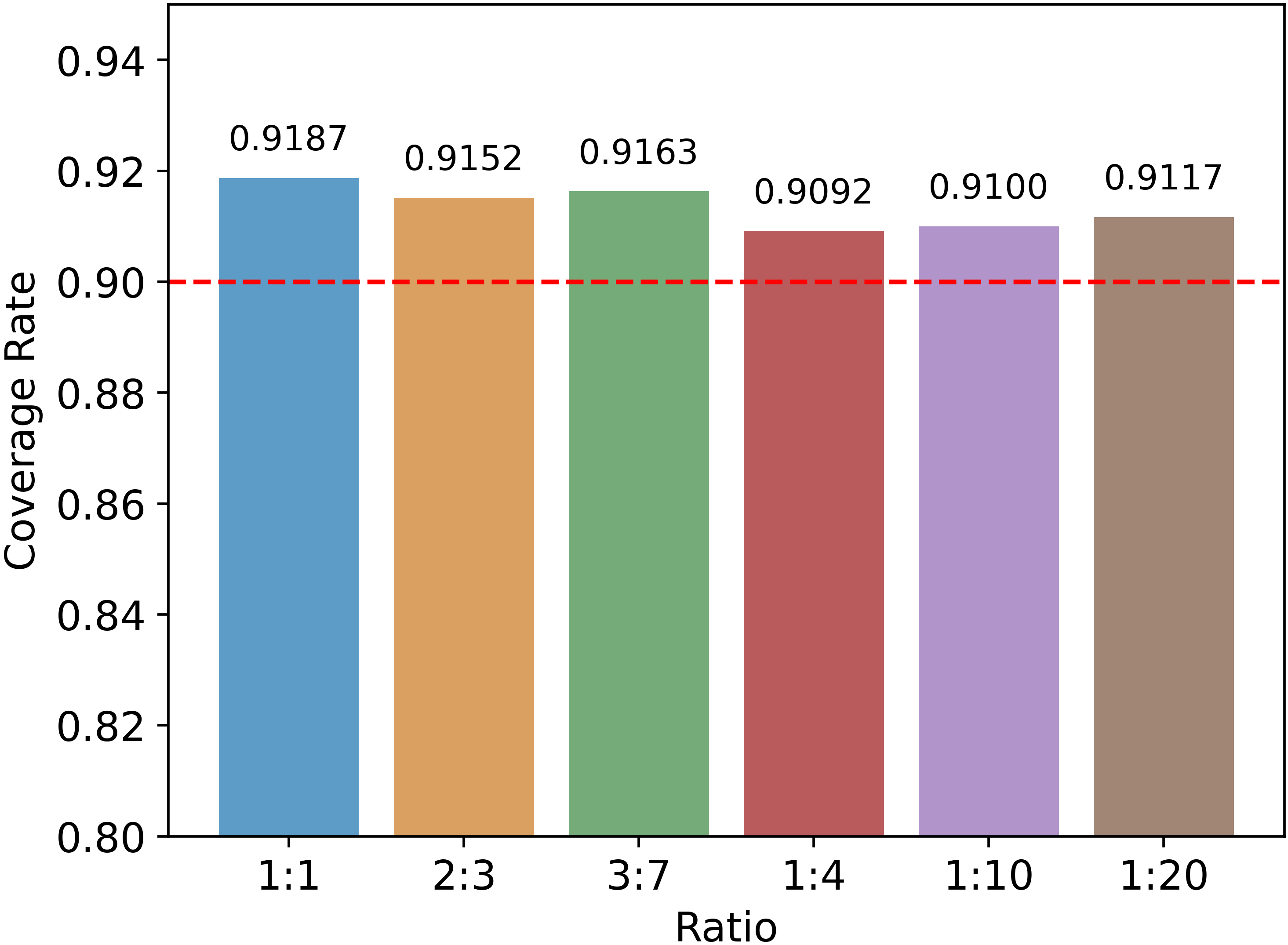}}
\caption{The average coverage rate across 4 datasets at different ratios between the calibration and test set utilizing the LLaMA-3-8B-Instruct model. The red dashed line indicates the lower bound at 0.9 (i.e., $\alpha=0.1$).}
\label{fig: multi ratio cp}
\end{figure}

As described in Section~\ref{sec: conformal prediction}, conformal prediction assumes a calibration set for the threshold $\hat{q}$. 
In our prior analysis, We divide the dataset into the calibration and test set at a fixed ratio of 1:10. 
Here, we investigate the correctness coverage rate at different ratios of size between the calibration and test set, and present the results in Figure~\ref{fig: multi ratio cp}. 
Despite various ratios of set size, we can always obtain a strict lower bound of the coverage rate by constructing prediction sets based on our devised conformal uncertainty criterion. 
This indicates the potential impacts of our method for robust guarantees in real-world open-ended NLG applications. 

%% file: tables/coverage_rate.tex
\begin{table}[t!]
\centering
\caption{The results of correctness coverage rate ($\%$) on 7 LLMs with various sizes across 4 open-ended NLG datasets. The user-specified error rate $\alpha$ is set to 0.1.}
\adjustbox{max width=\linewidth}{
    \begin{tabular}{c|cccc} 
        \toprule
        LLMs & TriviaQA & CoQA & MedQA & MedMCQA\\
        \midrule
        
        LLaMA-2-7B-Chat & 91.00 & 93.37 & 100.00 & 91.32\\
        Mistral-7B-Instruct-v0.3 & 90.83 & 91.87 & 90.70 & 90.39\\
        
        LLaMA-3-8B-Instruct & 94.27 & 90.73 & 90.46 & 93.17\\
        LLaMA-2-13B-Chat & 91.68 & 91.63 & 91.72 & 92.45\\
        Vicuna-13B-v1.5 & 90.19 & 92.68 & 90.25 & 92.13\\
        LLaMA-3-70B-Instruct & 92.18 & 90.95 & 93.70 & 92.48\\
        GPT-3.5-turbo  & 93.14 & 91.66 & 91.78 & 90.36\\

        \bottomrule
    \end{tabular}
}
\label{tb: cr}
\end{table}

%% file: tables/prediction_set_size.tex
\begin{table}[t!]
\centering
\caption{The average prediction set size on 7 LLMs with various sizes across 4 open-ended NLG datasets. The user-specified error rate $\alpha$ is set to 0.1.}
\adjustbox{max width=\linewidth}{
    \begin{tabular}{c|cccc} 
        \toprule
        LLMs & TriviaQA & CoQA & MedQA & MedMCQA\\
        \midrule
        
        LLaMA-2-7B-Chat & 2.28 & 2.26 & 4.28 & 3.07\\
        Mistral-7B-Instruct-v0.3 & 2.24 & 2.49 & 4.20 & 3.26\\
        LLaMA-3-8B-Instruct & 2.34 & 2.45 & 2.68 & 2.60\\
        LLaMA-2-13B-Chat & 2.19 & 2.28 & 3.40 & 2.73\\
        Vicuna-13B-v1.5 & 2.26 & 2.47 & 3.29 & 2.98\\
        LLaMA-3-70B-Instruct & 1.03 & 1.71 & 2.15 & 1.60\\
        GPT-3.5-turbo & 1.96 & 2.13 & 2.49 & 2.02\\

        \bottomrule
    \end{tabular}
}
\label{tb: prediction set size}
\end{table}

%% file: tables/accuracy_calibration.tex
\begin{table}[t!]
\centering
\caption{The enhancement of model accuracy ($\%$) after conducting selective prediction within the calibrated prediction sets based on the black-box uncertainty measure, utilizing sentence similarity as the criterion for correctness evaluation under the threshold of 0.7. }
\adjustbox{max width=\linewidth}{
    \begin{tabular}{c|c|c|c} 
        \toprule
        Dataset & LLMs & Original & Calibrated\\
        \midrule
        
        \multirow{5}{*}{TriviaQA} & LLaMA-2-7B-Chat & 68.43 & 70.77\\
        \multirow{5}{*}{} & Mistral-7B-Instruct-v0.3 & 79.04 & 81.45\\
        \multirow{5}{*}{} & LLaMA-3-8B-Instruct & 79.36 & 80.00\\
        \multirow{5}{*}{} & Vicuna-13B-v1.5 & 78.40 & 78.80\\
        \multirow{5}{*}{} & LLaMA-2-13B-Chat & 76.70 & 78.13\\

        \midrule

        \multirow{5}{*}{CoQA} & LLaMA-2-7B-Chat & 73.00 & 75.53\\
        \multirow{5}{*}{} & Mistral-7B-Instruct-v0.3 & 78.25 & 80.80\\
        \multirow{5}{*}{} & LLaMA-3-8B-Instruct & 72.93 & 74.67\\
        \multirow{5}{*}{} & Vicuna-13B-v1.5 & 76.17 & 78.43\\
        \multirow{5}{*}{} & LLaMA-2-13B-Chat & 80.00 & 81.23\\

        \midrule

        \multirow{5}{*}{MedQA} & LLaMA-2-7B-Chat & 37.88 & 40.80\\
        \multirow{5}{*}{} & Mistral-7B-Instruct-v0.3 & 38.65 & 43.90\\
        \multirow{5}{*}{} & LLaMA-3-8B-Instruct & 66.29 & 70.59\\
        \multirow{5}{*}{} & Vicuna-13B-v1.5 & 44.42 & 46.78\\
        \multirow{5}{*}{} & LLaMA-2-13B-Chat & 42.07 & 46.15\\
        
        \bottomrule
    \end{tabular}
}
\label{tb: acc}
\end{table}

%% file: section/conclusion.tex
\section{Conclusion}
In this work, we introduce \textit{ConU} tailored for black-box UQ in open-ended NLG tasks. 
Relying on CP which can transform any heuristic approximation into a statistically rigorous uncertainty notion, we develop a robust conformal uncertainty criterion to provide reliable guarantees of correctness coverage under various user-specified error rates. 
We achieve strict control of the coverage rate across 7 practical LLMs on 4 free-from NLG datasets. 
Furthermore, the small average uncertainty set size underscores the efficiency of our methods. 
Utilizing these calibrated prediction sets, we perform selective prediction and obtain remarkable improvements in model accuracy. 
We envisage that our conformal uncertainty criterion can provide new strategies for principled UQ in open-ended NLG tasks. 

%% file: section/appendix.tex
\section{Proof of the Coverage Property}
This is the explanation of validity for the conformal uncertainty criterion introduced in Section~\ref{sec: conformal prediction}. 
We reproduce the derivation here for completeness. 
Let us break down the overall implementation into the following five steps: 

\textbf{Black-box Uncertainty Measure.} We first conduct semantic clustering within the $M$ candidate generations and obtain $K$ non-repeated semantics for each sample. 
Since generations in the $k$-th cluster share the equivalent meaning, we denote any one generation in the $k$-th cluster as $\hat{y}_k^{(i)}$. 
Then we rely on self-consistency and define the uncertainty score of each candidate generation as $\mathcal{U}\left( \hat{y}_m^{(i)} \right)$ as described in Eq.~\eqref{eq: uncertainty score}. 

\textbf{NS Definition.} For each calibration sample, we select the generation that (1) first shares the equivalent semantics with the reference answer and (2) then exhibits the highest semantic similarity to the reference answer, and then define the NS as its uncertainty score calculated following Eq.~\eqref{eq: uncertainty score}. 
The first condition is to tightly couple the NS with correctness and the second is to facilitate generation selection in test samples. 
The NS of the $i$-th calibration data $r_i$ is described as Eq.~\eqref{eq: NS}. 

\textbf{Conformal Uncertainty Criterion.} We calculate the $\frac{\left \lceil \left ( N+1\right )\left ( 1-\alpha \right )\right \rceil}{N}$ quantile of the NSs for all fresh calibration data to develop our conformal uncertainty criterion (i.e., the uncertainty threshold $\hat{q}$) based on the user-specified error rate $\alpha$. 
As described in Eq.~\ref{eq: q_hat}, $\hat{q}={r}_{\left \lceil \left ( N+1\right )\left ( 1-\alpha \right )\right \rceil}$. 

\textbf{Construction of Prediction Sets.} For each test data, we construct a prediction set following Eq.~\eqref{eq: prediction set}. 
Since the generation that is semantically equivalent to $\hat{y}_i^{(test)}$ and shares the highest semantic similarity to $\hat{y}_i^{(test)}$ in $\left\{ \hat{y}_m^{(test)} \right\}_{m=1}^M$ is itself, we can obtain $r\left( x_{test}, \hat{y}_j^{(test)} \right) = \mathcal{U}\left( \hat{y}_j^{(test)} \right)$. 
Then we calibrate the prediction set by selecting generations, of which the uncertainty satisfies the conformal uncertainty criterion closely linked with correctness. 

\textbf{Correctness Coverage Guarantees.} Considering the assumption that there is at least one correct answer in $\left\{ \hat{y}_m^{(test)} \right\}_{m=1}^M$, we can conclude that the event $\left\{ y_{test}^* \in  \mathcal{P}\left(x_{test}\right) \right\}$ is equivalent to $\left\{ r_{test}=r\left( x_{test}, y_{test}^* \right) \leq \hat{q} \right\}$. 
Since $\left(x_1, y_1^*\right)$, ..., $\left(x_N, y_N^*\right)$, $\left(x_{test}, y_{test}^*\right)$ are exchangeable, we have $P\left( r_{test} \leq r_i \right)=\frac{i}{N+1}$. 
Ultimately, we achieve rigorous guarantees of the correctness coverage rate on test samples as described as Eq.~\eqref{eq: lower bound}. 

\section{Validity of Assumption (1)}\label{sec: assumption1}
We assume that at least one acceptable response is sampled into the candidate set for each test data point. 
For each calibration data point, we sample multiple generations from the output space, denoted as $\mathcal{C}_{m}\left( X_i \right) = \left\{ \hat{Y}_j^{(i)} \right\}_{j=1}^{m}$. 
Then, we define the loss of miscoverage by the candidate set as 
\begin{equation}
    l\left( \mathcal{C}_{m}\left( X_i \right), Y_i^* \right)=\mathbf{1}\left\{ Y_i^* \notin \mathcal{C}_{m}\left( X_i \right) \right\},
\end{equation}
and the loss is non-increasing in $m$.

We set $A_N\left(m\right) = \displaystyle\sum_{i=1}^{N} l\left( \mathcal{C}_{m}\left( X_i \right), Y_i^* \right)$. 
Given that $l\left( \mathcal{C}_{m}\left( X_{test} \right), Y_{test}^* \right) \in \left\{0,1\right\}$, we obtain 
\begin{equation}
\begin{split}
    &A_{N+1}\left( m \right) \\
    &=  
    \displaystyle\sum_{i=1}^{N+1} l\left( \mathcal{C}_{m}\left( X_i \right), Y_i^* \right)\\
    &=A_N\left(m\right) + l\left( \mathcal{C}_{m}\left( X_{test} \right), Y_{test}^* \right)\\
    &\in \left\{A_N\left(m\right),A_N\left(m\right)+1\right\}.
\end{split}
\end{equation}

By the exchangeability of $N$ calibration data points and the test data point, we have $l_{test} \backsim \mathrm{Uniform} \left( \left\{ l_1, \cdots, l_{N}, l_{test} \right\} \right)$, where $l_{i}$ is the abbreviation for $l\left( \mathcal{C}_{m}\left( X_{i} \right), Y_{i}^* \right)$~\cite{angelopoulosconformal}. 
Then, we have
\begin{equation}
\begin{split}
    &\mathbb{E}\left[ l\left( \mathcal{C}_{m}\left( X_{test} \right), Y_{test}^* \right) \right]\\ &= \frac{A_{N+1}\left( m \right)}{N+1}\\
    &\in \left\{\frac{A_{N}\left( m \right)}{N+1},\frac{A_{N}\left( m \right)+1}{N+1}\right\}.
\end{split}
\end{equation}

Since we have demanded that at least one acceptable response is sampled into the candidate set for each calibration data (i.e., $A_{N}\left( m \right)=0$), we obtain $\mathbb{E}\left[ l\left( \mathcal{C}_{m}\left( X_{test} \right), Y_{test}^* \right) \right] \in \left\{0,\frac{1}{N+1}\right\}$ and Assumption (1) holds in this case. 

\section{Implementation Details}
\label{sec: implementation details}
\subsection{Baselines}
\label{sec: ap baselines}
We compare \textit{ConU} with 8 baseline measures.   
\textit{PE} is defined as the entropy over the whole generation and \textit{LNPE} is the length normalized \textit{PE}. 
\textit{SE} tackles the issue of semantic equivalence by gathering generations sharing the same meaning into semantic clusters and calculating cluster-wise entropy. 
\textit{SAR} solves the issue of generative inequality and allocates more attention to key tokens and sentences. 
\textit{LS} measures the average sentence similarity among sampled responses. 
\textit{NumSet} employs the number of semantic sets (equivalence classes) as a reflection of uncertainty. 
\textit{Deg} and \textit{Ecc} treat each generation as one node,  calculate the symmetric normalized graph Laplacian, and respectively utilize the degree matrix and the average distance from the center as the uncertainty measures. 

We do not compare the two recent approaches that adapt CP for correctness coverage in open-ended NLG tasks for several reasons: (1) Conformal language modeling~\cite{quach2023conformal} relies on the white-box model likelihoods information, which is impractical for recent LLMs served via API without logit access; 
(2) LofreeCP~\cite{su2024api} is susceptible to different settings of datasets and models, and cannot consistently guarantee the correctness coverage rate; 
(3) Our conformal uncertainty criterion achieves strict control of the correctness coverage rate under various user-specified error rates, model settings, and datasets, first linking black-box UQ with rigorous guarantees of correctness coverage, which meets the requirement for general NLG applications.

\subsection{Datasets}
\label{sec: ap datasets}
CoQA~\cite{reddy2019coqa} is a large-scale conversational QA dataset with more than 127k question-answer pairs equipped with contextual information. 
TriviaQA~\cite{joshi2017triviaqa} is a reading comprehension dataset with over 650k question-answer pairs. 
MedQA~\cite{jin2021disease} is a medical MCQA dataset collected from professional medical board exams. 
MedMCQA~\cite{pal2022medmcqa} is a large-scale MCQA dataset for practical medical entrance exam questions. 
For the evaluation of UQ, we randomly select 3,000 samples from each dataset. 
For the verification of correctness coverage guarantees, we utilize the development set (7,983 questions) of CoQA and full validation sets of MedQA and MedMCQA. 
For TriviaQA, we utilize the same 3,000 samples in UQ evaluations. 

For CoQA, we utilize the contextual information combined with the question as the prompt. 
For TriviaQA and MedMCQA, we randomly select 5 question-answer pairs as a fixed few-shot template and combine it with the current question. 
For MedQA, we employ 3 question-answer pairs. 

\section{Robustness of Conformal Uncertainty Criterion}
\label{sec: ap correctness coverage guarantees}
\begin{figure}[!h]
\centerline{\includegraphics[width=\columnwidth]{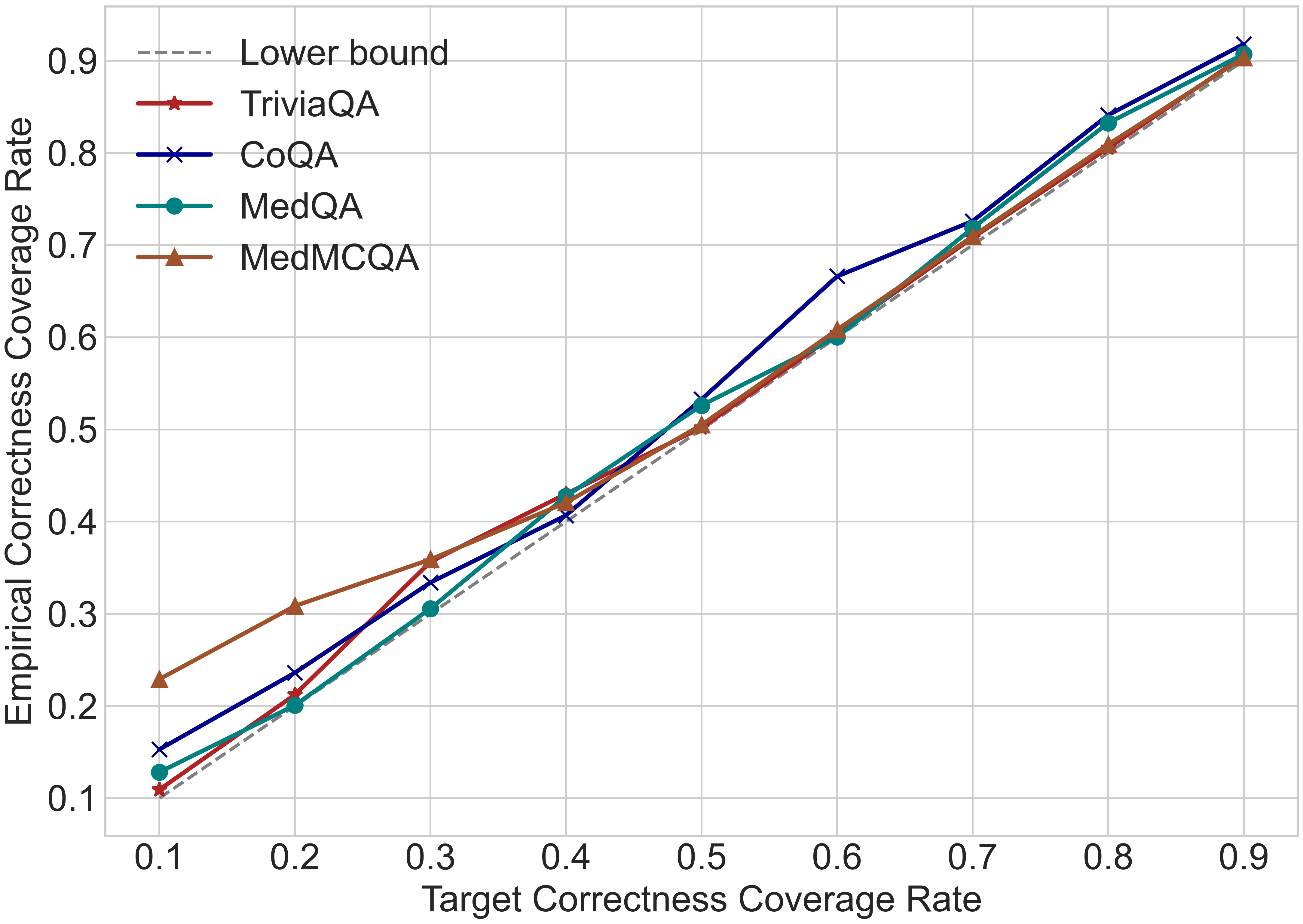}}
\caption{Target vs. empirical correctness coverage rate.\\We test the 4 datasets utilizing the Mistral-7B-Instruct-v0.3 model as the generator.}\label{fig: coverage mistral}
\end{figure}

\begin{figure}[!h]
\centerline{\includegraphics[width=\columnwidth]{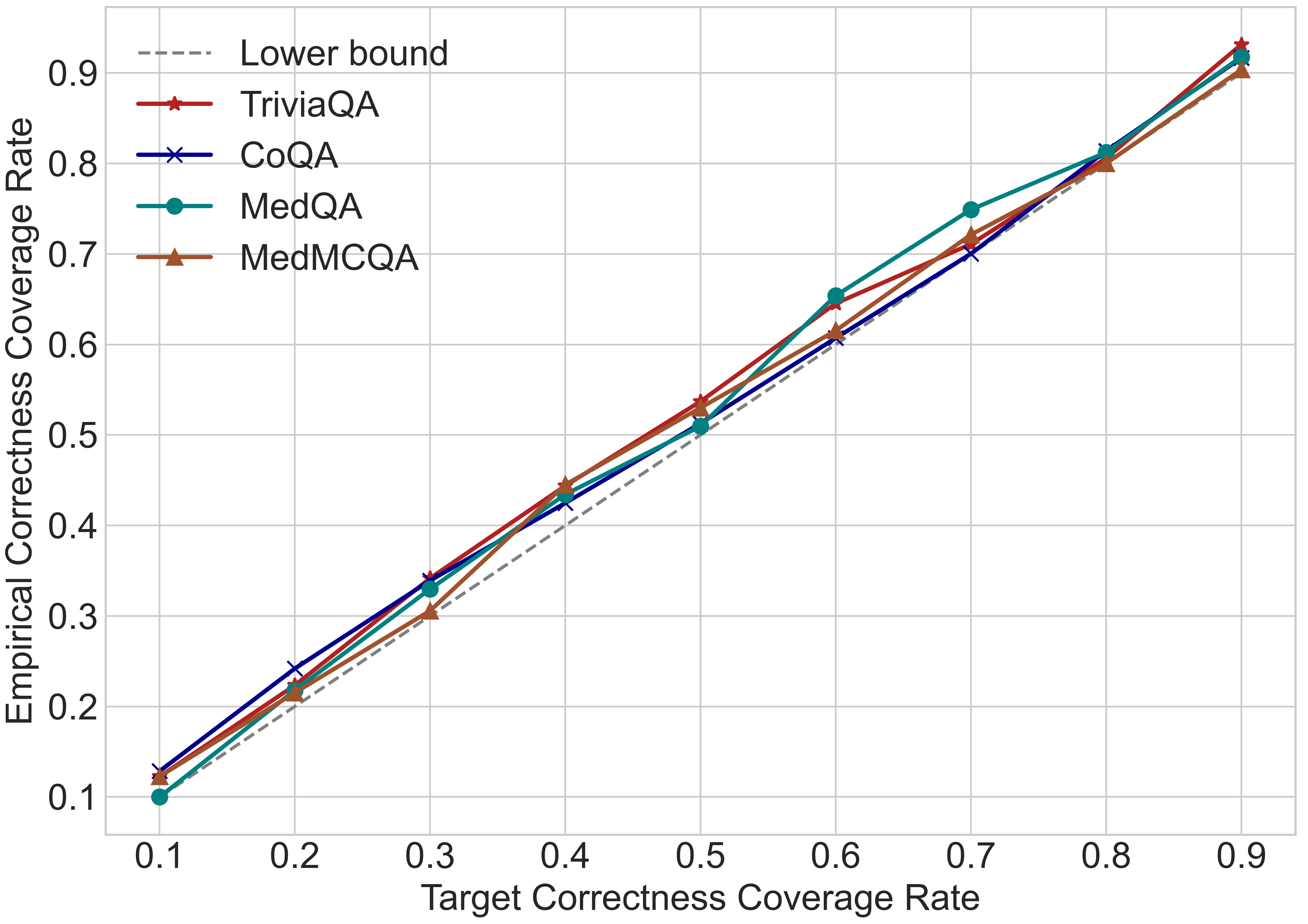}}
\caption{Target vs. empirical correctness coverage rate. We test the 4 datasets utilizing the GPT-3.5-turbo model as the generator.}\label{fig: coverage gpt}
\end{figure}

We verify the correctness coverage guarantees on the other 6 LLMs across 4 datasets. 
As demonstrated in Figures~\ref{fig: coverage mistral} \~ ~\ref{fig: coverage gpt}, we achieve rigorous control of coverage rate under various user-specified error rates despite different model settings or datasets. 
We also report the results of the correctness coverage rate under two strict error rates of 0.05 and 0.01. 
Table~\ref{tb: cr 0.05} and Table~\ref{tb: cr 0.01} indicate the robustness of our conformal uncertainty criterion. 

\begin{figure}[!h]
\centerline{\includegraphics[width=\columnwidth]{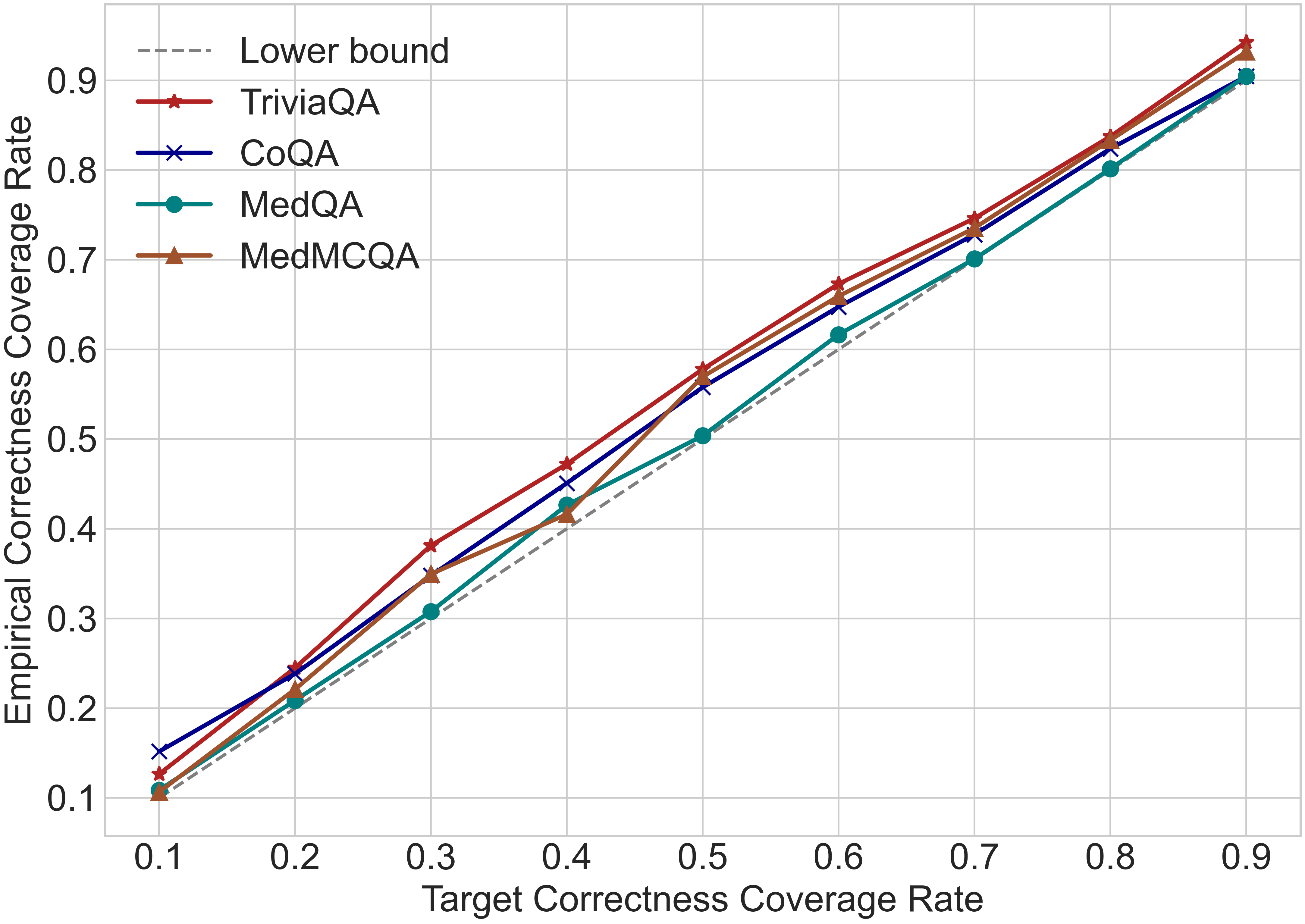}}
\caption{Target vs. empirical correctness coverage rate.\\We test the 4 datasets utilizing the LLaMA-3-8B-Instruct model as the generator.}\label{fig: coverage llama8b}
\end{figure}

\begin{figure}[!h]
\centerline{\includegraphics[width=\columnwidth]{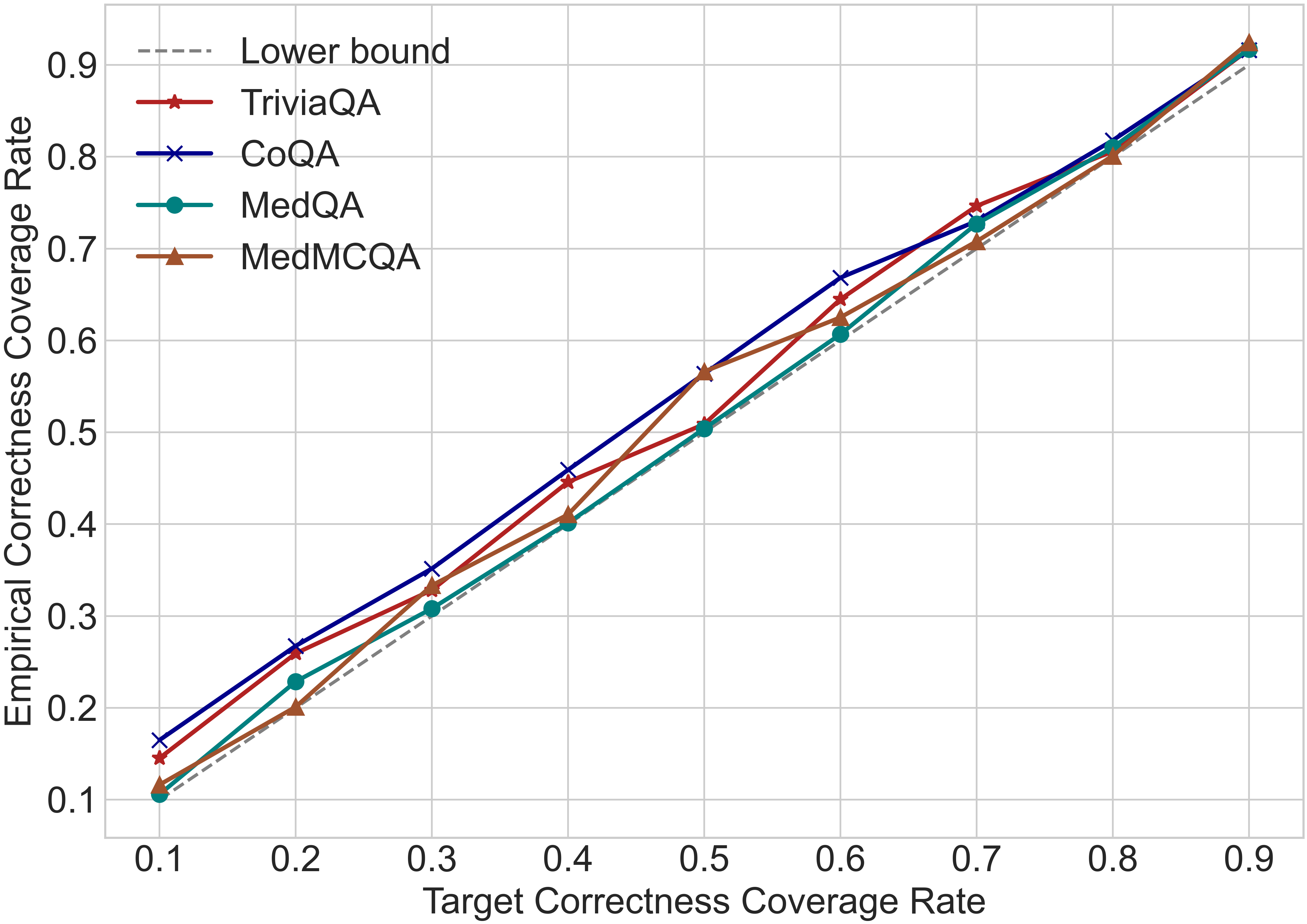}}
\caption{Target vs. empirical correctness coverage rate. We test the 4 datasets utilizing the LLaMA-2-13B-Chat model as the generator.}\label{fig: coverage llama13b}
\end{figure}

\begin{figure}[!h]
\centerline{\includegraphics[width=\columnwidth]{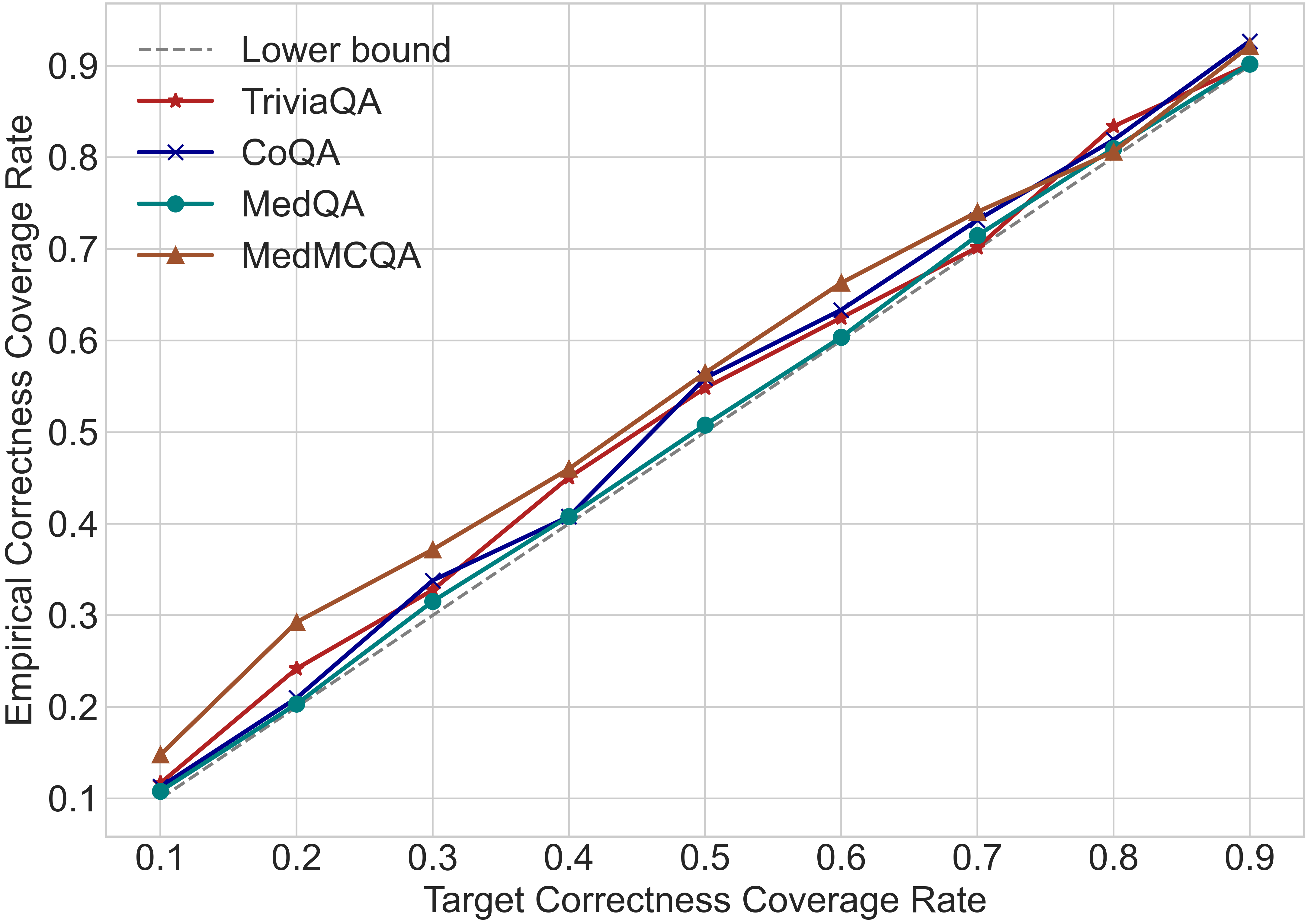}}
\caption{Target vs. empirical correctness coverage rate. We test the 4 datasets utilizing the Vicuna-13B-v1.5 model as the generator.}\label{fig: coverage vicuna}
\end{figure}

\begin{figure}[!h]
\centerline{\includegraphics[width=\columnwidth]{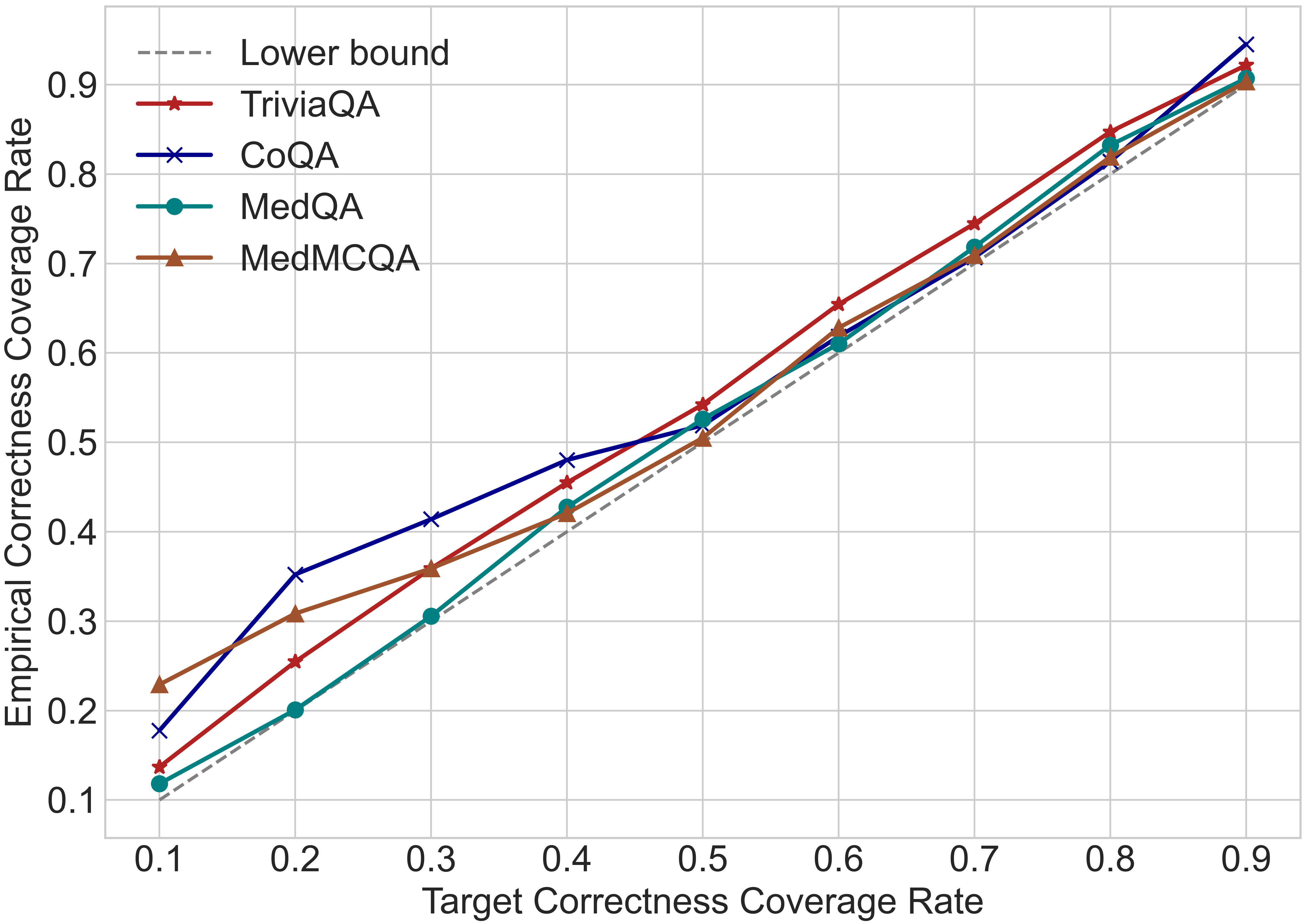}}
\caption{Target vs. empirical correctness coverage rate.\\We test the 4 datasets utilizing the LLaMA-3-70B-Instruct model as the generator.}\label{fig: coverage llama70b}
\end{figure}

\input{tables/coverage_rate_0.05}

\input{tables/coverage_rate_0.01}

%% file: tables/coverage_rate_0.05.tex
\begin{table}[h!]
\centering
\caption{The results of correctness coverage rate ($\%$) on 7 LLMs across 4 open-ended NLG datasets. The user-accepted error rate $\alpha$ is strictly set to 0.05.}
\adjustbox{max width=\linewidth}{
    \begin{tabular}{c|cccc} 
        \toprule
        LLMs & TriviaQA & CoQA & MedQA & MedMCQA\\
        \midrule
        
        LLaMA-2-7B-Chat & 95.26 & 96.45 & 100.00 & 95.99\\
        Mistral-7B-Instruct-v0.3 & 95.01 & 95.72 & 95.79 & 95.12\\
        LLaMA-3-8B-Instruct & 98.17 & 95.23 & 95.78 & 98.38\\
        LLaMA-2-13B-Chat & 95.04 & 96.96 & 95.15 & 96.59\\
        Vicuna-13B-v1.5 & 97.28 & 95.33 & 95.51 & 97.29\\
        LLaMA-3-70B-Instruct & 95.38 & 95.33 & 95.51 & 97.29\\
        GPT-3.5-turbo & 97.02 & 97.60 & 95.62 & 95.19\\

        \bottomrule
    \end{tabular}
}
\label{tb: cr 0.05}
\end{table}

%% file: tables/coverage_rate_0.01.tex
\begin{table}[h!]
\centering
\caption{The results of correctness coverage rate ($\%$) on 7 LLMs across 4 open-ended NLG datasets. The user-accepted error rate $\alpha$ is strictly set to 0.01.}
\adjustbox{max width=\linewidth}{
    \begin{tabular}{c|cccc} 
        \toprule
        LLMs & TriviaQA & CoQA & MedQA & MedMCQA\\
        \midrule
        
        LLaMA-2-7B-Chat & 99.93 & 99.83 & 100.00 & 99.14\\
        Mistral-7B-Instruct-v0.3 & 99.38 & 99.27 & 99.15 & 99.81\\
        LLaMA-3-8B-Instruct & 99.79 & 99.53 & 100.00 & 99.76\\
        LLaMA-2-13B-Chat & 99.06 & 99.13 & 99.51 & 99.48\\
        Vicuna-13B-v1.5 & 99.52 & 100.00 & 99.94 & 100.00\\
        LLaMA-3-70B-Instruct & 99.84 & 99.75 & 99.15 & 99.82\\
        GPT-3.5-turbo & 99.17 & 99.82 & 99.51 & 99.95\\

        \bottomrule
    \end{tabular}
}
\label{tb: cr 0.01}
\end{table}